\documentclass{article}

\PassOptionsToPackage{numbers, compress}{natbib}
\usepackage[preprint]{neurips_2024}
\usepackage[numbers]{natbib} 

\usepackage[utf8]{inputenc} % allow utf-8 input
\usepackage[T1]{fontenc}    % use 8-bit T1 fonts
\usepackage{hyperref}       % hyperlinks
\usepackage{url}            % simple URL typesetting
\usepackage{booktabs}       % professional-quality tables
\usepackage{amsfonts}       % blackboard math symbols
\usepackage{nicefrac}       % compact symbols for 1/2, etc.
\usepackage{microtype}      % microtypography
\usepackage{xcolor}         % colors
\usepackage{amsthm,amsmath,amssymb}
\usepackage{mathrsfs}
\usepackage{graphicx}
\usepackage{subfigure}
\usepackage{multirow}
\usepackage{caption}

\title{Continual Cross-Modal Generalization}

\author{%
  Yan Xia$^{*}$ \\
  Zhejiang University\\
  \texttt{xiayan.zju@gmail.com} \\
  \And
  Hai Huang\thanks{Equal Contribution} \\
  Zhejiang University\\
  \texttt{haihuangcode@outlook.com} \\
  \And
  Minghui Fang\\
  Zhejiang University\\
  \texttt{minghuifang@zju.edu.cn} \\
  \And
  Zhou Zhao\thanks{Corresponding Author}\\
  Zhejiang University\\
  \texttt{zhouzhao@zju.edu.cn} \\
}

\begin{document}

\maketitle

\begin{abstract}
  Cross-modal generalization aims to learn a shared discrete representation space from multimodal pairs, enabling knowledge transfer across unannotated modalities. However, achieving a unified representation for all modality pairs requires extensive paired data, which is often impractical. Inspired by the availability of abundant bimodal data (e.g., in ImageBind), we explore a continual learning approach that incrementally maps new modalities into a shared discrete codebook via a mediator modality. We propose the Continual Mixture of Experts Adapter (CMoE-Adapter) to project diverse modalities into a unified space while preserving prior knowledge. To align semantics across stages, we introduce a Pseudo-Modality Replay (PMR) mechanism with a dynamically expanding codebook, enabling the model to adaptively incorporate new modalities using learned ones as guidance. Extensive experiments on image-text, audio-text, video-text, and speech-text show that our method achieves strong performance on various cross-modal generalization tasks. Code is provided in the supplementary material.
\end{abstract}

\section{Introduction}
With the explosive growth of multimodal data, many efforts \cite{girdhar2023imagebind, radford2021learning, liang2022mind, zhao2022towards} have been made toward mapping these diverse modalities into a shared semantic space to reduce the semantic gap between paired modalities. Works based on contrastive learning, i.e., models like CLIP \cite{radford2021learning} in the image-text domain and ImageBind \cite{girdhar2023imagebind}, which connects multiple modalities centered around images, learn inter-modal alignment from large-scale paired data. Additionally, many studies employ modality-agnostic encoders \cite{akbari2021vatt, you2022learning} to map different modalities into the same semantic space. While these approaches bring different modalities closer in a shared semantic space, a significant domain gap remains.

Recently, some research \cite{zhao2022towards, liu2021cross, duan2022multi} has focused on leveraging explicit vector quantization (VQ)~\cite{van2017neural, ji2024wavtokenizer} or prototypes to map features from different modalities with the same semantics to identical discrete variables, achieving notable progress. Compared to implicit alignment methods, using discrete variables allows for better aggregation of similar features, thereby enabling faster convergence of inter-modal alignment. However, these methods typically compress features from different modalities into a single vector before mapping, making fine-grained alignment challenging.

To address the aforementioned issues, the Cross-Modal Generalization \cite{xia2024achieving} (CMG) task has been proposed. It aims to map features from different modalities with the same semantics into a common codebook by learning fine-grained, unified multimodal representations from extensive paired pre-training data. During the downstream phase, the model can transfer knowledge from one modality and generalize to other unseen modalities. Their approach achieves fine-grained alignment and demonstrates significant performance in multiple downstream cross-modal generalization tasks. However, the successful training of their model heavily relies on a substantial amount of paired multimodal data. For instance, achieving unified representation for more than three modalities necessitates acquiring paired data for all corresponding modalities, which is often rare and limits the advancement of research in unified multimodal representation.

Inspired by ImageBind \cite{girdhar2023imagebind}, pairwise combinations of modalities are typically abundant and easily accessible, where one modality serves as an intermediary to connect other disparate modalities. In this paper, we primarily investigate how to leverage the continual learning paradigm to incrementally incorporate new modalities into the existing pre-trained data based on CMG framework \cite{xia2024achieving}. 
However, there are two intrinsic limitations existing in the previous CMG framework when applied to this scenario:
\textbf{1.} CMG do not support incremental learning, making it impossible to add new modal data to an already trained common discrete space.
\textbf{2.} Due to the diverse origins of paired modal data, newly introduced paired modalities may contain entirely different semantic information compared to previous modalities. Directly employing methods like CMG could erroneously map these new semantic categories into the existing discrete variables. Therefore, the model needs a dynamically growing discrete dictionary that can expand with the introduction of new modalities.

Thus in this paper, we aim to gradually extend the unified representation from two modalities to multiple modalities, ensuring that the alignment of original modalities remains intact while mapping new modalities into the existing semantic space.
Addressing the abovementioned challenges, we propose the following innovations to mitigate these issues:
\textbf{1.} To enable the model to map different modalities into a unified semantic space, we propose a \textbf{Continual Mixture of Experts Adapter} (CMoE-Adapter) module. First, we employ a multimodal universal adapter structure, which includes a multimodal shared Common Layer and modality-specific Specific Layers. Different modalities activate their corresponding Specific Layers. However, a single Adapter structure is insufficient for encoding capabilities in the face of continuously expanding modalities. Therefore, we transform it into a Mixture of Experts (MoE) Adapter architecture, which leverages different combinations of experts to handle various scenarios, thereby enhancing the model's encoding capacity. To ensure that the Encoder's encoding ability for previous modalities remains unchanged when introducing new modalities for training, we incorporate an adaptive Elastic Weight Consolidation (EWC) loss,  choosing important parameters within the MoE Adapter.
\textbf{2.} To address the second issue, we propose a \textbf{Pseudo-Modality Replay} mechanism (PMR). Initially, we introduce a dynamic expansion codebook, mapping unknown semantic information from new modalities to newly added dictionary codes. Then we construct a sequence as a pseudo-modality by selecting the nearest code from the expanded codebook based on the intermediary modality. The pseudo-modality allows the model to leverage knowledge learned from previous data as a teacher, thereby allowing the new codebook to encompass previously learned discrete semantic features while adapting to new modality features. In summary, our contributions are threefold:
\begin{itemize}
    \item We extend the previous cross-modal generalization task to a continual cross-modal generalization task, effectively leveraging the vast amount of pairwise data available on the internet to train a unified representation space across multiple modalities.
    \item We propose \textbf{COMET} (\textbf{CO}ntinuing \textbf{M}ultimodal unified r\textbf{E}presen\textbf{T}ation learning) framework, which contains the CMoE-Adapter and PMR module, which allow the model to expand its training data in a continual learning manner, mapping new modalities into the previously learned common discrete space.
    \item We pre-train on video-text, audio-text, image-text, and speech-text datasets to obtain a unified representation space for various modalities. Extensive experiments on downstream tasks such as audio-video, speech-video, and video-text cross-modal generalization demonstrate the effectiveness of our model.
\end{itemize}

\begin{figure*}[t] 
 \center{\includegraphics[width=14cm]  {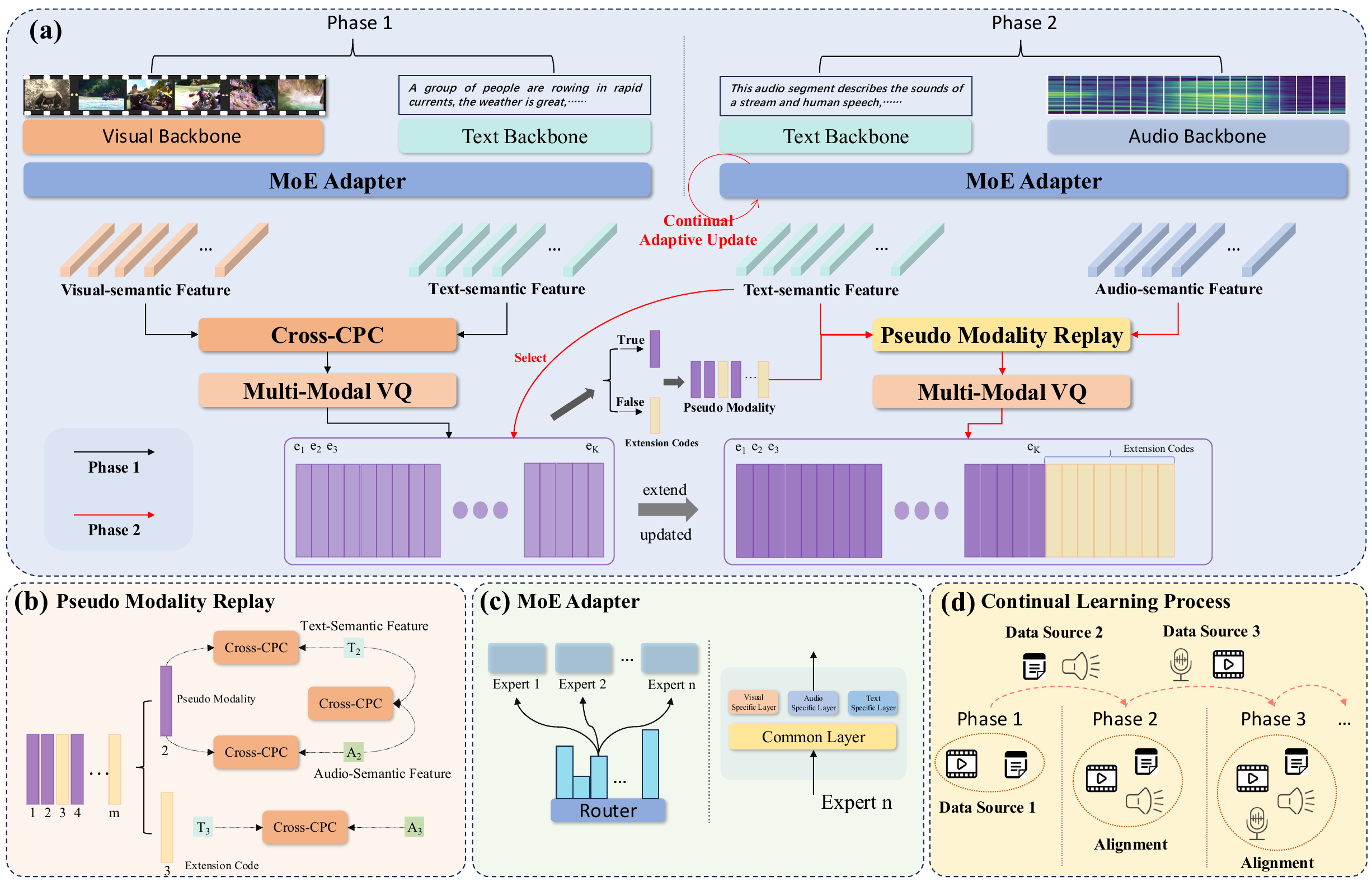}} 
 \caption{\label{fig2} The overview of our proposed continual unified multimodal representation framework, we use two stages as an example. We replicate the codebook obtained from the previous phase for use in the new phase, extend it, and continuously update it during subsequent training. However, during the PMR process, the codes acquired from the previous phase remain unchanged.} 
\end{figure*}
\vspace{-1.0em}

\section{Backgrounds of CMG}
\textbf{Cross Modal Generalization Task:}
Given a multi-modal dataset $\mathbb{X} = \{(\mathbf{x}_{i}^{A}, \mathbf{x}_{i}^{B}, \mathbf{x}_{i}^{C}...)\}_{i=1}^{N}$ with N instances across modalities A, B, C, etc., the Cross Modal Generalization (CMG) task seeks to unify these modalities into a common discrete space during pre-training. This alignment allows for shared discrete latent codes corresponding to identical semantics. In subsequent tasks, knowledge from a single annotated modality (e.g., A) can be extrapolated to unannotated ones (e.g., B and C) via the pre-trained common space, achieving zero-shot cross-modal generalization ability.
%首先介绍本任务的定义，以及相较于传统的CMG有哪些不同

\textbf{Baseline Details:} 
For the paired modalities $\{(\mathbf{x}_{i}^{a}, \mathbf{x}_{i}^{b})\}_{i=1}^{N}$, CMG aims to extract universal features $\mathbf{z}_{i}^{a}, \mathbf{z}_{i}^{b}$ using semantic encoders ${\Phi}^{a}$ and ${\Phi}^{b}$, and modality-specific features $\mathbf{\bar{z}}_{i}^{a}, \mathbf{\bar{z}}_{i}^{b}$ using encoders ${\Psi}^{a}$ and ${\Psi}^{b}$. They then quantize the semantic features into discrete codes using a vector quantization (VQ) operation, shared across both modalities. Finally, these codes are combined with the modality-specific features to reconstruct the original input features.

\textbf{Dual Cross-modal Information Disentangling (DCID):} CMG presents DCID module that operates on two fronts: minimizing mutual information (MI) for intra-modality features using the CLUB approach and maximizing MI for inter-modality semantic features with Cross-CPC.

\textbf{CLUB-based MI Minimization:} Employing a temporal adaptation of CLUB, they aim to reduce the correlation between semantic ($\mathbf{z}_{i}^{m}$) and modality-specific features ($\bar{\mathbf{z}}_{i}^{m}$), alternating optimization between the approximation and the main networks during pre-training:
\begin{equation}
    I_{\text{vCLUB}}(\mathbf{x}; \mathbf{y}) :=  \mathbb{E}_{p(\mathbf{x},\mathbf{y})} [\log q_\theta(\mathbf{y}|\mathbf{x})]  -\mathbb{E}_{p(\mathbf{x})}\mathbb{E}_{p(\mathbf{y})} [\log q_\theta(\mathbf{y}| \mathbf{x})],
\end{equation}

\textbf{Cross-CPC for MI Maximization:} Extending the principle of Contrastive Predictive Coding (CPC) for cross-modal predictions, they utilize the predictive nature of autoregressive models to maximize information shared across modalities, considering human-like inference abilities. For this, they employ unidirectional LSTM to summarize past information up to a time $t$ for modalities A and B. Predicting the k-th future step in modality B from A (and vice versa), they optimize the InfoNCE loss with a linear projection matrix $W_{k}^{m}$ for each step:
\begin{equation}
    L_{cpc}^{a2b} = -\frac{1}{K}\sum_{k=1}^{K}log\Big[\frac{exp(\mathbf{z}_{t+k}^{b}W_{k}^{a}\mathbf{c}_{t}^{a})}{\sum_{\mathbf{z}_{j} \in Z_{b}}exp(\mathbf{z}_{j}^{b}W_{k}^{a}\mathbf{c}_{t}^{a})}\Big];
    \quad L_{cpc}^{b2a} = -\frac{1}{K}\sum_{k=1}^{K}log\Big[\frac{exp(\mathbf{z}_{t+k}^{a}W_{k}^{b}\mathbf{c}_{t}^{b})}{\sum_{\mathbf{z}_{j} \in Z_{a}}exp(\mathbf{z}_{j}^{a}W_{k}^{b}\mathbf{c}_{t}^{b})}\Big],
\end{equation}

\textbf{Multi-modal Exponential Moving Average (MM-EMA):} CMG introduced the MM-EMA model for fine-grained cross-modal alignment, utilizing a teacher-student dynamic for iterative updates during quantization. Cross-attention mechanisms help map semantic features across modalities. For instance, with modality A as the query and modality B as the key and value, they derive a vector $\mathbf{r}^{b}_{i}$ that correlates with $\mathbf{z}_{i}^{a}$ and retains characteristics of modality B. This intermediary facilitates EMA alignment. 
Quantized to a code vector $\mathbf{e}_{i}$, they combine semantic vectors from both modalities, calculating updated counts $N_i^{(t)}$ and code vector volumes $\mathbf{o}_i^{(t)}$ as follows, $\gamma$ is the decay factor:
\begin{align}
    N_i^{(t)} &= \gamma N_i^{(t-1)} + (1-\gamma)[n_i^{a(t)} + n_i^{b(t)}]\;\;\; \mathbf{e}_i^{(t)} = \mathbf{o}_i^{(t)} / N_i^{(t)}\\ \nonumber
    \mathbf{o}_i^{(t)} = \gamma &\mathbf{o}_i^{(t-1)} + (1-\gamma)\Big[\sum_{j=1}^{n_i^{a(t)}}\frac{\mathbf{z}_{i,j}^{a(t)}+\mathbf{r}_{i,j}^{b(t)}}{2} + \sum_{j=1}^{n_i^{b(t)}}\frac{\mathbf{z}_{i,j}^{b(t)}+\mathbf{r}_{i,j}^{a(t)}}{2}\Big],
\end{align}

\section{Continual Cross Modal Generalization}

\subsection{Task Definition}
% First, we give the definition of our proposed continuous cross modal generalization task. 
Given few paired multi-modal datasets $\mathbb{X} = \{(\mathbf{x}_{i}^{A}, \mathbf{x}_{i}^{B})\}_{i=1}^{N_{1}}$, $\mathbb{Y} = \{(\mathbf{y}_{i}^{A}, \mathbf{y}_{i}^{C})\}_{i=1}^{N_{2}}$, $\mathbb{Z} = \{(\mathbf{z}_{i}^{A}, \mathbf{z}_{i}^{D})\}_{i=1}^{N_{3}}$..., where A, B, C represent different modalities, $N_{1}$, $N_{2}$, $N_{3}$ represent instance numbers of different datasets. In these datasets, an intermediate modality exists where the contained semantic information exhibits a degree of partial overlap. In this task, we plan to train a unified multi-modal representation space in a phased manner. During the pre-training stage, we initially train the unified representation of two modalities from dataset $\mathbb{X}$. Subsequently, using modality A as a bridge, we train the unified representation of three modalities (A, B, and C) in the second step with dataset $\mathbb{Y}$, followed by the unified representation of four modalities (A, B, C, and D) in the third step, and so forth. Then in downstream tasks, the pre-trained unified multi-modal representation space enables zero-shot knowledge transfer and generalization across these four modalities.

In this paper, we build upon the CMG approach by introducing the Continual MoE Adapter and the Pseudo Modality Reply module. These additions endow the model with the capability to incrementally expand the unified multi-modal representation space, using a shared modality as a bridge. The following sections will provide a detailed introduction to these two modules.

\subsection{Continual MoE Adapter}
\textbf{Expert Design} Although the CMG approach can link paired multi-modal features, it struggles to directly map newly introduced, unknown modalities into the same semantic space as previous modalities in a continual learning scenario. Therefore, we need to employ intermediate modalities as bridges between different datasets to map unpaired modalities into a shared latent space. Inspired by prior work, we utilize a common layer as a shared mapping layer for different modalities, enabling the decoupled semantic features of various modalities to be mapped together. However, a single common layer may lead to forgetting the feature mapping capabilities of previous modalities when learning new ones. To address this, we restructured the architecture to combine a common layer with multiple modality-specific layers. The common layer contains a single linear component, sharing information across modalities, while the specific layers encode modality-unique information, the specific layer for each modality also contains a single linear component. This ensures that during continual learning, the common layer can gradually learn the features shared across multiple modalities. The whole architecture is an expert, named $E(x)$. 
%再强调我们在每个阶段只对数据集中有的模态激活对应的specific layer

\textbf{Mixture of Experts} Single experts often struggle to handle complex multi-modal scenarios effectively. Therefore, we extend this approach to a Mixture of Experts (MoE) framework, which integrates knowledge from various domains through the combination of different experts. Previous studies have demonstrated that this structure is more adept at adapting to continual learning environments. We employ a linear layer followed by a SoftMax function to serve as a router, selecting the appropriate combination of experts. The forward propagation process of our MoE block can be mathematically expressed as follows:
\begin{equation}
    MoE(x) = \sum_{i=1}^{O}G_{i}(x)E_{i}(x),\quad G(x) = SoftMax(W_{g}(x)),
\end{equation}
where O is the expert number, $W_{g}$ is the gate linear function. To mitigate the issue of a few experts being disproportionately activated during MoE training, we employ a load balancing loss as a constraint.
Let $G \in \mathbb{R}^{B \times O}$ be the gate outputs for a batch of size B, where O is the number of experts. Each element $G_{ij}$ represents the gating value for the i-th input and the j-th expert. The load for each expert can be computed by summing the gating values over the batch: $ L_j = \sum_{i=1}^B G_{ij}$, where \( L_j \) is the load for the \( j \)-th expert. The ideal load  I for each expert is: $\text{I} = \frac{B}{U}$, where U is the number of the experts. Finally, the load balancing loss can be defined to penalize deviations from the ideal load:  
\begin{equation}
    \mathcal{L}_{\text{gate}} = \frac{1}{U} \sum_{j=1}^U \left( \frac{L_j}{\text{I}} - 1 \right)^2
\end{equation}

\textbf{Adding Elastic Weight Consolidation (EWC) Loss:} To prevent the model parameters from forgetting previously learned features when encoding newly introduced modalities and new semantic categories during the next training phases, we apply EWC loss to both the common layers and the modality-specific layers that have been activated within each expert of MoE. This approach selectively penalizes changes to the most important parameters of our model, thus protecting previously learned knowledge while accommodating new information. 

To be detailed, in our continual learning framework, we denote the parameters of the common layer and the activated specific layer within each expert of the MoE structure as $\theta$, the EWC loss is formulated to constrain the update of these parameters by considering the importance of each parameter to previous tasks. The importance is quantified by the Fisher Information Matrix (FIM), FF, which is computed on the previous tasks' data. Given the optimal parameters $\theta^*$ obtained after training on the previous tasks, the EWC loss for our linear layer is defined as:

\begin{equation}
\mathcal{L}_{\text{EWC}}(\theta) = \sum_i \frac{\lambda}{2} F_i (\theta_i - \theta_i^*)^2, \quad F_i = \mathbb{E}_{x \sim \mathcal{D}} \left[ \left( \frac{\partial \log p(y|x,\theta^*)}{\partial \theta_i} \right)^2 \right]
\end{equation}
where \( \theta_i \) represents the current value of the \(i\)-th parameter, \( \theta_i^* \) is its value after training on previous tasks, \( F_i \) is the corresponding entry in the Fisher Information Matrix, and \( \lambda \) is a hyperparameter that controls the strength of the regularization, $\mathcal{D}$ is the data distribution of the previous tasks.

\subsection{Pseudo Modality Reply}

As training progresses, paired modalities from different datasets are introduced, often containing semantic categories distinct from those in the previously trained data. The previous CMG approach fixes the number of codes in the codebook during training. Applying this fixed codebook to new scenarios can result in different semantics being erroneously mapped to the same discrete variables, thereby disrupting the pre-trained codes. Therefore, it is crucial to consider how to map new semantic features to new codes in the subsequent training phases, while allowing the old semantic features to update the pre-trained codebook.

In this section, we design a Pseudo Modality Reply (PMR) module. We first design a dynamic expansion codebook. The unknown semantic information from new data is mapped to newly added dictionary sequences. Simultaneously, the codebook obtained from the previous training phase is utilized as a Teacher, enabling the new codebook to encompass previously learned discrete semantic features while also accommodating new modality features. This approach ultimately learns a mature unified discrete representation space across multiple modalities.

During the initial training phase, our approach is identical to traditional Cross Modal Generalization (CMG). In subsequent phases, when new datasets are introduced, we first replicate the previously trained codebook $V_{1}$ of size \(K1\) to the current phase's codebook $V_{2}$ and add \(K2\) new codes, resulting in a codebook of size \((K1 + K2)\). Assuming modality B is the common intermediate modality across both phases' datasets, we construct a pseudo-modality by selecting the nearest codes from the previous codebook based on modality B. This pseudo-modality represents the semantic features of the previous dataset and serves as a teacher to guide the convergence of the current codebook. We use the semantic feature of modality B to select the nearest codes from the previous codebook $V_{1}$: If a code matching the semantics of the current B-modal sequence can be found in V1, the corresponding code is retrieved from V1; otherwise, it is selected from the newly added K2 codes (to obtaining a discrete sequence of semantic features of the same length as the current modalities). We refer to the code sequence mapped from the semantics of the B-modal sequence as the pseudo-modal sequence. This sequence is used for Cross-CPC \cite{xia2024achieving} computation with the B and C modal sequences during the current training phase. In this process, we assign $e^{-6}$ weights to the codes from K2 within the pseudo-modal sequence, primarily utilizing the codes retrieved from V1 to compute the convergence of the B and C modalities.

\subsection{Training and Downstream Tasks}
The full objective of our continual pre-training framework is the combination of the objection functions: $L = L_{recon} + L_{commit}+ L_{cpc} +  L_{cmcm} + L_{MI} + L_{gate}$, where the six losses are those utilized in the previous CMG \cite{xia2024achieving} model. $L_{EWC}$ is specially optimized for MoE adapters.
We divide the entire continual learning pre-training process into three stages, with each stage utilizing a paired modality of data. The training data from the previous stage is not accessible in the subsequent stages. Upon completion of the pre-training, the unified representation encoder and codebook obtained are employed for cross-modal generalization in downstream tasks. In these downstream tasks, we train the model on labeled data from one modality and then transfer the trained model to other unlabeled modalities to evaluate its performance.
%%%%%%%%%%%%%%%%%%%%%%%%%%%%%%%%%%%%%%%%%%%%%%%%%%%%%%%%%%%%

\section{Experiments}
\label{experiment}
% \textbf{Pre-train:} We use VATEX~\cite{} as video-text pair pretrain data, 通过爬取得到有效视频-文本对xx个. use AudioCaps~\cite{} as audio-text pair pretrain data, 过滤得到有效音频-文本对xx个. use flickr30k~\cite{} as image-text pair pretrain data , 过滤得到有效图片-文本对xx个, use LibriTTS-train-clean-100~\cite{} as speech-text pair pretrain data, 过滤得到有效音语音-文本对xx个.

% \textbf{Downstream:} we evaluate the pre-trained models on several downstream tasks using different datasets. Audio-Video:{\bf Cross-modal event classification on AVE dataset:}~\cite{avel} training on one modality (\eg video) and evaluating on another (\eg audio). {\bf Cross-modal event localization on AVVP dataset:}~\cite{tian2020unified} localizing events in one modality and transferring to the other. {\bf Cross-dataset localization/classification:} training on classification in AVE and evaluating localization in AVVP, transferring across datasets.
% Speech-Text:
% {\bf Cross-modal video segmentation on Ref-Youtube-VOS dataset:}~\cite{} segmenting video based on an speech or text query, Ref-Youtube-VOS没有speech数据，这部分是从AVOS~\cite{}取的，取Ref-Youtube-VOS和AVOS的且集合, use ReferFormer~\cite{} as downstream modal.
% {\bf Zero-shot Retrieval:}
% Video-Text: MSVD~\cite{},采用类似MSRVTT~\cite{}的处理，只取MSVD test子集中每个视频的第一条caption.
% image-audio: flickr sound~\cite{}
% Audio-text: clotho~\cite{}
% image-text: MS-COCO~\cite{}

\subsection{Pre-train}
\textbf{Video-Text Pairs:} We use VATEX~\cite{wang2019vatex} as video-text pair pre-training data, filtering to obtain effective 30251 video-text pairs through web crawling. \textbf{Audio-Text Pairs:} We use AudioCaps~\cite{kim2019audiocaps} as audio-text pair pre-training data, filtering to obtain effective 33055 audio-text pairs. \textbf{Image-Text Pairs:} We use Flickr30k~\cite{young2014image} as image-text pair pre-training data, filtering to obtain effective 31783 image-text pairs. \textbf{Speech-Text Pairs:} We use LibriTTS-train-clean-100~\cite{zen2019libritts} as speech-text pair pre-training data, filtering to obtain effective 31862 speech-text pairs.

\subsection{Downstream}
The downstream tasks involve various paired data, including video-audio, video-text, audio-text, image-text, speech-text, and image-audio pairings, as detailed in Appendix~\ref{sec:appendix_downstream}.

\subsection{Tasks Setting}
The downstream test datasets can be adjusted based on the datasets selected for pre-training and are divided into two parts: seen data pairs (data directly paired during pre-training) and unseen data pairs (data not directly paired during pre-training). For example, in VT-AT-IT pre-training, VT (video-text) and AT (audio-text) are seen data pairs, while VA (video-audio) and IA (image-audio) are unseen data pairs.

\subsection{Implementation Details}
\textbf{Baselines:} We compare our results with those of TURN~\cite{zhao2022towards}, CMCM~\cite{liu2021cross}, and DCID~\cite{xia2024achieving} by replicating their experiments and enhancing them with the Comet module, which improves specific aspects of the model's performance. The pairing order of audio-text, video-text, image-text, and speech-text during pre-training can be freely arranged, and additional combinations like video-audio pairs can also be introduced into the training process. The primary experiments are demonstrated using a three-stage model consisting of video-text (VT), audio-text (AT), and image-text (IT) stages. For the Cross-modal Video Segmentation on the Ref-Youtube-VOS Dataset~\cite{seo2020urvos} task, a model sequence of video-text (VT), speech-text (ST), and audio-text (AT) is exemplified.

\textbf{Metrics:} For the AVE~\cite{avel}, precision $(\%)$ is used as the metric; the f1-score $(\%)$ is utilized for assessing the AVVP~\cite{tian2020unified} and the AVE to AVVP generalization task. The region similarity(J) and the contour accuracy(F) are used for the Ref-Youtube-VOS~\cite{seo2020urvos} dataset, and recall is utilized for zero-shot retrieval.

\begin{table}[t]
\caption{Results in video-audio pairs. $\dagger$ indicates pre-training in the order of AT-VT-IT and the other settings in the order of VT-AT-IT.}
\label{tab:unseen_va}
\centering % 将\centering移到表格外部
\resizebox{\textwidth}{!}{%
\begin{tabular}{cccccccccccccccc}
\toprule
\multirow{3}{*}{Method} & \multicolumn{7}{c}{Second Stage} & \multicolumn{1}{c}{} & \multicolumn{7}{c}{Third Stage} \\  
\cmidrule(lr){2-8} \cmidrule(lr){10-16}
& \multicolumn{2}{c}{AVE} & \multicolumn{2}{c}{AVVP} & \multicolumn{2}{c}{AVE $\rightarrow$ AVVP} & \multicolumn{1}{c}{Avg.} & & \multicolumn{2}{c}{AVE} & \multicolumn{2}{c}{AVVP} & \multicolumn{2}{c}{AVE $\rightarrow$ AVVP} & \multicolumn{1}{c}{Avg.} \\
& V $\rightarrow$ A & A $\rightarrow$ V & V $\rightarrow$ A & A $\rightarrow$ V & V $\rightarrow$ A & A $\rightarrow$ V & & & V $\rightarrow$ A & A $\rightarrow$ V & V $\rightarrow$ A & A $\rightarrow$ V & V $\rightarrow$ A & A $\rightarrow$ V \\
\midrule
TURN~\cite{zhao2022towards} &  16.62 & 15.36 & 16.51 & 17.12 & 20.15 & 20.38 & 17.69 & & 15.57 & 19.35 & 21.9 & 19.4 & 24.16 & 14.07 & 19.07 \\
CMCM~\cite{liu2021cross} & 19.05 & 18.15 & 22.28 & 18.16 & 24.16 & 22.33 & 20.68 & & 20.32 & \textbf{23.72} & 20.03 & 18.54 & 28.11 & 27.86 & 23.09 \\
DCID~\cite{xia2024achieving} & 16.56 & 15.45 & 25.73 & 21.66 & 22.16 & 23.74 & 20.88 & & 15.27 & 17.76 & 25.05 & \textbf{23.87} & 26.1 & \textbf{33.92} & 23.66 \\
TURN~\cite{zhao2022towards}+COMET & 24.46 & \textbf{21.17} & 17.84 & 17.7 & 27.3 & 17.11 & 20.93 & & \textbf{20.45} & 23.59 & 20.03 & 18.54 & 18.84 & 28.4 & 21.64 \\
CMCM~\cite{liu2021cross}+COMET & 21.29 & 20.03 & 24.78 & 23.61 & 26.41 & \textbf{27.38} & 23.92 & & 20.42 & 22.67 & 25.69 & 17.08 & 31.29 & 28.23 & 24.23 \\
DCID~\cite{xia2024achieving}+COMET & \textbf{25.30} & 19.97 & \textbf{26.06} & 20.46 & \textbf{31.16} & 23.46 & \textbf{24.41} & & 17.88 & 22.01 & \textbf{29.59} & 23.06 & \textbf{34.36} & 28.95 & \textbf{25.97} \\
\midrule
DCID~\cite{xia2024achieving} $\dagger$ & 15.90 & 15.12 & 25.84 & 22.47 & \textbf{29.01} & \textbf{35.68} & 24.01& & 13.72 & 18.98 & 24.45 & \textbf{23.19} & 24.81 & \textbf{34.03} & 23.19 \\
DCID~\cite{xia2024achieving}+COMET $\dagger$ & \textbf{17.55} & \textbf{24.95} & \textbf{26.99} & \textbf{24.94} & 27.15 & 32.36 & \textbf{25.66} & & \textbf{13.77} & \textbf{20.57} & \textbf{28.28} & 22.28 & \textbf{30.08} & 29.25 & \textbf{24.04} \\
\bottomrule
\end{tabular}%
}
\end{table}

\subsection{Performance on unseen data pairs}

As we can see in Table \ref{tab:unseen_va} and Table \ref{tab:unseen_IA} in Appendix \ref{sec:appendix_seen}, it is evident that COMET significantly enhances the performance of other models on unseen data pairs, encompassing tasks such as classification and localization of video-audio pairs, as well as zero-shot retrieval of image-audio pairs. Specifically, for video-audio related tasks, the four models trained under the VT-AT-IT paradigm exhibit an average improvement of over 3\% in the Second Stage and approximately 2\% in the Third Stage. Furthermore, when altering the pre-training order, the results remain consistently significant, indicating that COMET is robust to variations in training sequence and emphasizes the unified representation of multimodal data.

\subsection{Performance on seen data pairs}

We also altered the training sequence and conducted experiments on seen data pairs, as shown in Tables \ref{tab:seen_vt}, Table \ref{tab:seen_at} and Table \ref{tab:seen_it}, finding that COMET consistently enhances baseline performance across various modality-related tasks. Specifically, in the video-text pairs task, COMET improves performance at all three stages, even with the changed training sequence. Furthermore, COMET maintains its advantage in audio-text and image-text tasks as well.
\begin{table}[t]
\caption{Retrieval results in video-text pairs. $\dagger$ indicates pre-training in the order of AT-VT-IT and the other settings in the order of VT-AT-IT.}
\label{tab:seen_vt}
\resizebox{\textwidth}{!}{%
\begin{tabular}{cccccccccccccccccccccccc}
\toprule
\centering
\multirow{3}{*}{Method} & \multicolumn{7}{c}{First Stage} & & \multicolumn{7}{c}{Second Stage} & & \multicolumn{7}{c}{Third Stage} \\
\cmidrule{2-8} \cmidrule{10-16} \cmidrule{18-24} 
& \multicolumn{3}{c}{T $\rightarrow$ V} & \multicolumn{3}{c}{V $\rightarrow$ T} & \multirow{2}{*}{Avg.} &  & \multicolumn{3}{c}{T $\rightarrow$ V} & \multicolumn{3}{c}{V $\rightarrow$ T} & \multirow{2}{*}{Avg.} & & \multicolumn{3}{c}{T $\rightarrow$ V} & \multicolumn{3}{c}{V $\rightarrow$ T} & \multirow{2}{*}{Avg.}\\
& R@1 & R@5 & R10 & R@1 & R@5 & R10 & & & R@1 & R@5 & R10 & R@1 & R@5 & R10 & & & R@1 & R@5 & R10 & R@1 & R@5 & R10  \\
\midrule
TURN~\cite{zhao2022towards} & 2.69 & 9.1 & 15.22 & 2.54 & 9.4 & 15.37 & 9.05 & & 0.6 & 2.24 & 4.48 & 0.3 & 2.39 & 4.03 & 2.34 & & 2.99 & 9.1 & 13.13 & 2.39 & 7.46 & 13.43 & 8.08 \\
CMCM~\cite{liu2021cross} & 2.69 & 10 & 16.42 & 3.13 & 10.9 & 17.91 & 10.17 & & 0.6 & 2.54 & 5.07 & \textbf{0.6} & 1.94 & 4.33 & 2.51 & & 2.99 & \textbf{10.3} & 14.48 & 1.94 & 9.25 & 14.93 & 8.98 \\
DCID~\cite{xia2024achieving} & \textbf{3.58} & 11.49 & 18.66 & 2.69 & 10.45 & 17.76 & 10.77 & & 0.75 & 2.99 & 5.37 & 0.3 & 1.79 & 3.88 & 2.51 & & 2.24 & 9.4 & 16.87 & \textbf{3.13} & 8.51 & 14.48 & 9.11 \\\

TURN~\cite{zhao2022towards}+COMET & 2.99 & 10.45 & 16.87 & \textbf{3.58} & 11.04 & 16.87 & 10.3 & & 0.6 & 2.24 & 4.48 & 0.3 & 2.39 & 4.03 & 2.34 & & 2.39 & 9.55 & 15.97 & 1.64 & 8.21 & 14.48 & 8.71 \\
CMCM~\cite{liu2021cross}+COMET & 2.54 & \textbf{12.84} & 19.4 & \textbf{3.58} & 11.19 & 18.81 & 11.39 & & 0.45 & 3.43 & 6.87 & 0.3 & 2.09 & 5.22 & 3.06 & & \textbf{3.28} & 9.7 & 15.67 & \textbf{3.13} & \textbf{10.3} & 14.18 & 9.38 \\
DCID~\cite{xia2024achieving}+COMET & 2.84 & 12.09 & \textbf{19.7} & 2.84 & \textbf{11.64} & \textbf{19.85} & \textbf{11.49} & & \textbf{1.04} & \textbf{4.03} & \textbf{7.31} & \textbf{0.6} & \textbf{4.03} & \textbf{7.01} & \textbf{4.01} & & 2.99 & 9.7 & \textbf{17.01} & 2.99 & 9.85 & \textbf{15.82} & \textbf{9.73} \\
\midrule
DCID~\cite{xia2024achieving} $\dagger$ & - & - & - & - & - & - & - & & \textbf{3.58} & 11.49 & 18.96 & 3.13 & 12.24 & \textbf{20.45} & 11.64 & & 2.39 & 9.25 & 14.03 & \textbf{2.39} & 8.66 & 13.73 & 8.41 \\
DCID~\cite{xia2024achieving}+COMET $\dagger$ & - & - & - & - & - & - & - & & 3.43 & \textbf{12.84} & \textbf{20.3} & \textbf{3.73} & \textbf{12.69} & 18.81 & \textbf{11.96} & & \textbf{3.88} & \textbf{10.9} & \textbf{16.87} & 2.09 & \textbf{10.15} & \textbf{15.52} & \textbf{9.91}\\
\bottomrule
\end{tabular}%
}
\end{table}

\begin{table}[h]
\caption{Retrieval results in audio-text pairs. $\dagger$ indicates pre-training in the order of AT-VT-IT and the other settings in the order of VT-AT-IT.}
\label{tab:seen_at}
\resizebox{\textwidth}{!}{%
\begin{tabular}{cccccccccccccccccccccccc}
\toprule
\centering
\multirow{3}{*}{Method} & \multicolumn{7}{c}{First Stage} & & \multicolumn{7}{c}{Second Stage} & & \multicolumn{7}{c}{Third Stage} \\
\cmidrule{2-8} \cmidrule{10-16} \cmidrule{18-24} 
& \multicolumn{3}{c}{T $\rightarrow$ A} & \multicolumn{3}{c}{A $\rightarrow$ T} & \multirow{2}{*}{Avg.} & & \multicolumn{3}{c}{T $\rightarrow$ A} & \multicolumn{3}{c}{A $\rightarrow$ T} & \multirow{2}{*}{Avg.} &  & \multicolumn{3}{c}{T $\rightarrow$ A} & \multicolumn{3}{c}{A $\rightarrow$ T} & \multirow{2}{*}{Avg.} \\
& R@1 & R@5 & R10 & R@1 & R@5 & R10 & & &  R@1 & R@5 & R10 & R@1 & R@5 & R10 & & & R@1 & R@5 & R10 & R@1 & R@5 & R10 & \\
\midrule
TURN~\cite{zhao2022towards} & - & - & - & - & - & - & - & & 6.22 & 22.68 & 36.46 & 5.74 & 18.66 & 29.19 & 19.83  & & 2.2 & 12.63 & 21.15 & 1.91 & 10.05 & 18.66 & 11.1 \\
CMCM~\cite{liu2021cross} & - & - & - & - & - & - & - & &  8.52 & 24.59 & 38.47 & 7.18 & 25.84 & 35.41 & 23.33 & & 3.06 & 13.68 & 24.31 & 0.96 & 11.96 & 18.18 & 12.03 \\
DCID~\cite{xia2024achieving} & - & - & - & - & - & - & - & & \textbf{9.86} & \textbf{27.08} & 39.81 & 5.74 & 24.4 & 35.41 & 23.72  && 3.06 & 12.73 & 20.67 & 2.39 & 8.13 & 16.27 & 10.54 \\
TURN~\cite{zhao2022towards}+COMET & - & - & - & - & - & - & - & &  8.04 & 25.36 & 38.56 & 7.18 & 20.57 & 33.01 & 22.12 & & 3.06 & 14.45 & \textbf{25.26} & 2.39 & 9.57 & 14.35 & 11.51 \\
CMCM~\cite{liu2021cross}+COMET & - & - & - & - & - & - & - & & 8.61 & 26.79 & \textbf{40.77} & 7.18 & 23.92 & 36.84 & 24.02 & & \textbf{4.02} & \textbf{15.5} & 23.64 & \textbf{3.83} & 11.48 & \textbf{21.53} & \textbf{13.33} \\
DCID~\cite{xia2024achieving}+COMET & - & - & - & - & - & - & - & &  \textbf{9.86} & 25.84 & 38.09 & \textbf{7.66} & \textbf{27.27} & \textbf{37.32} & \textbf{24.34}& & 3.16   &13.97 & \textbf{25.26} & 2.87 & \textbf{12.92} & \textbf{21.53} & 13.28 \\

\midrule
DCID~\cite{xia2024achieving} $\dagger$ &6.99 & \textbf{25.65} & 36.56 & 6.22 & 22.97 & 32.54 & 21.82 & &  3.06 & 10.62 & 18.66 & 2.39 & 8.61 & 14.83 & 9.69 & & \textbf{2.87} & 8.13 & 16.65 & 0.48 & 7.18 & 11.0 & 7.72 \\
DCID~\cite{xia2024achieving}+COMET $\dagger$ & \textbf{7.27} & 23.25 & \textbf{37.42} & \textbf{10.53} & \textbf{23.44} & \textbf{37.32} & \textbf{23.21} & &  \textbf{3.07} & \textbf{11.48} & \textbf{20.29} & \textbf{5.74} & \textbf{11.0} & \textbf{18.66} & \textbf{11.71} & & 2.68 & \textbf{11.1} & \textbf{17.8} & \textbf{2.39} & \textbf{8.61} & \textbf{13.4} & \textbf{9.33} \\
\bottomrule
\end{tabular}%
}
% \vspace{-0.5cm}
\end{table}

\begin{table}[h]
\caption{Retrieval results in image-text pairs. $\dagger$ indicates pre-training in the order of AT-VT-IT and the other settings in the order of VT-AT-IT.}
\label{tab:seen_it}
\centering
\resizebox{0.6\textwidth}{!}{%
\begin{tabular}{cccccccc}
\toprule
\centering
\multirow{3}{*}{Method} & \multicolumn{7}{c}{Third Stage} \\
\cmidrule{2-8}
& \multicolumn{3}{c}{T $\rightarrow$ I} & \multicolumn{3}{c}{I $\rightarrow$ T} & \multirow{2}{*}{Avg.} \\
& R@1 & R@5 & R10 & R@1 & R@5 & R10 &\\
\midrule
TURN~\cite{zhao2022towards} & 2.7 & 12 & 19.1 & 3.4 & 11.1 & 18.1 & 11.06 \\
CMCM~\cite{liu2021cross} & 2.6 & 11.5 & 19.4 & \textbf{3.7} & \textbf{13.4} & 20.5 & 11.85 \\
DCID~\cite{xia2024achieving} & 3.6 & 11.5 & 20.4 & \textbf{3.7} & \textbf{13.4} & 21.5 & 12.35 \\

TURN~\cite{zhao2022towards}+COMET & 3.3 & 12.1 & 18.9 & 3.1 & 13.1 & 20.5 & 11.83 \\
CMCM~\cite{liu2021cross}+COMET & \textbf{3.7} & 12.5 & 20.3 & 3.3 & 12.1 & 20.5 & 12.06 \\
DCID~\cite{xia2024achieving}+COMET & 3.5 & \textbf{13.5} & \textbf{21.1} & 3.2 & 11.8 & \textbf{23.2} & \textbf{12.72} \\

\midrule
DCID~\cite{xia2024achieving} $\dagger$ & 2.8 & 11.3 & 18.7 & \textbf{3.0} & 11.6 & 19.1 & 11.08 \\
DCID~\cite{xia2024achieving}+COMET $\dagger$ & \textbf{3.6} & \textbf{12.5} & \textbf{21.0} & \textbf{3.0} & \textbf{13.4} & \textbf{22.1} & \textbf{12.6} \\
\bottomrule
\end{tabular}%
}
\end{table}

Table \ref{tab:seen_st} presents the semantic segmentation results of the model on the Ref-Youtube-VOS dataset for the Text2Speech task after training on VT2ST2AT. In more complex semantic segmentation scenarios, Comet continues to enhance model performance. Furthermore, the inclusion of COMET consistently improves the performance of the DCID model in both the Second Stage and Third Stage.

\begin{table}[th]
\caption{Performance of semantic segmentation tasks for cross-modal generalization.((pre-trained on VT-ST-AT))}
\label{tab:seen_st}
\centering
\resizebox{0.6\textwidth}{!}{%
\begin{tabular}{cccccccc}
\toprule
\centering
\multirow{2}{*}{Method} & \multicolumn{3}{c}{Second Stage} & & \multicolumn{3}{c}{Third Stage}  \\   \cmidrule{2-4} \cmidrule{6-8}
& J  & F & J\&F & & J  & F & J\&F \\
\midrule
DCID~\cite{xia2024achieving} & 0.283 & 0.318 & 0.3005 & & 0.269 & 0.284 & 0.2765 \\
DCID~\cite{xia2024achieving}+COMET & \textbf{0.314} & \textbf{0.347} & \textbf{0.3305} & & \textbf{0.273} & \textbf{0.295} & \textbf{0.284} \\
\bottomrule
\end{tabular}%
}
\end{table}

\subsection{Ablation study}
\begin{table}[th]
\caption{Ablation studies of Video-Audio in the second and third training stage.(pre-trained on VT-AT-IT)}
\label{tab:ablation_va}
\centering % 将\centering移到表格外部
\resizebox{\textwidth}{!}{%
\begin{tabular}{cccccccccccccccccccc}
\toprule
\multirow{3}{*}{PM} & \multirow{3}{*}{MOE} & \multirow{3}{*}{$\mathcal{L}_{Gate}$} & \multirow{3}{*}{$\mathcal{L}_{EWC}$} & \multirow{3}{*}{$\mathcal{SL}$} & \multicolumn{7}{c}{Second Stage} & \multicolumn{1}{c}{} & \multicolumn{7}{c}{Third Stage} \\  
\cmidrule(lr){6-12} \cmidrule(lr){14-20}
& & & & & \multicolumn{2}{c}{AVE} & \multicolumn{2}{c}{AVVP} & \multicolumn{2}{c}{AVE $\rightarrow$ AVVP} & \multicolumn{1}{c}{Avg.} & & \multicolumn{2}{c}{AVE} & \multicolumn{2}{c}{AVVP} & \multicolumn{2}{c}{AVE $\rightarrow$ AVVP} & \multicolumn{1}{c}{Avg.} \\
& & & & & V $\rightarrow$ A & A $\rightarrow$ V & V $\rightarrow$ A & A $\rightarrow$ V & V $\rightarrow$ A & A $\rightarrow$ V & & & V $\rightarrow$ A & A $\rightarrow$ V & V $\rightarrow$ A & A $\rightarrow$ V & V $\rightarrow$ A & A $\rightarrow$ V \\
\midrule
- & $\checkmark$ & $\checkmark$ & $\checkmark$ & $\checkmark$ & \textbf{26.29} & 18.42 & 22.28 & 18.16 & 26.88 & 20.38 & 22.06 & & \textbf{23.01} & \textbf{22.28} & 26.21 & 22.35 & 28.31 & 22.33 & 24.08 \\
$\checkmark$ & - & - & - & - & 15.57 & \textbf{20.03} & 23.71 & 19.72 & 31.92 & \textbf{27.80} & 23.13 & & 19.91 & 19.04 & 28.03 & 20.62 & 32.13 & 29.03 & 24.79\\
$\checkmark$ & $\checkmark$ & - & $\checkmark$ & $\checkmark$ & 18.23 & 16.47 & 26.01 & \textbf{22.60} & \textbf{35.16} & 24.41 & 23.81 & & 16.11 & 19.28 & 28.58 & 23.57 & 31.29 & 30.52 & 24.89 \\
$\checkmark$ & $\checkmark$ & $\checkmark$ & - & $\checkmark$ & 23.89 & 18.92 & 25.84 & 20.72 & 31.73 & 21.40 & 23.75& & 16.62 & 18.42 & 28.03 & 21.81 & 33.39 & 31.05 & 24.88 \\
$\checkmark$ & $\checkmark$ & $\checkmark$ & $\checkmark$ & - & 18.52 & 19.46 & 23.53 & 21.62 & 30.96 & 26.6 & 23.45 & & 19.82 & 20.21 & 27.13 & 22.12 & 31.43 & 28.23 & 24.82 \\
- & - & - & - & - & 16.56 & 15.45 & 25.73 & 21.66 & 22.16 & 23.74 & 20.88 & & 15.27 & 17.75 & 25.05 & \textbf{23.87} & 26.10 & \textbf{33.92} & 23.66\\
$\checkmark$ & $\checkmark$ & $\checkmark$ & $\checkmark$ & $\checkmark$ & 25.31 & 19.97 & \textbf{26.06} & 20.46 & 31.16 & 23.46 & \textbf{24.41} & & 17.87 & 22.01 & \textbf{29.59} & 23.06 & \textbf{34.36} & 28.95 & \textbf{25.97} \\
\bottomrule
\end{tabular}%
}
\end{table}

\textbf{Effectiveness of each module:}
Based on the results presented in Table \ref{tab:ablation_va}, we observe that the removal of any single component results in a decline in model performance, thereby demonstrating the effectiveness of the proposed module.
Moreover, we observe that the removal of the Pseudo Modality (PM) module significantly impacts the model's performance. This is because the PM leverages the previously learned unified representations as a pseudo modality for alignment, effectively facilitating the relearning of prior knowledge and aiding the model in retaining this information.

Due to the integration of gate loss, EWC loss, and specific layers (SL) within the MOE architecture, the removal of the MOE also results in the elimination of these modules. As observed in Tables \ref{tab:ablation_va} and Table \ref{tab:ablation_ia} in Appendix \ref{sec:appendix_unseen}, this leads to a significant decline in average performance on the video-audio and image-audio tasks. These findings demonstrate that the MOE effectively aids COMET in extracting modality-specific information and fostering unified modality representation.

We retained the MoE model and selectively ablated the gate loss, EWC loss, and SL. Our findings indicate a performance drop compared to the full model; however, performance remained superior to the results obtained by completely removing the MoE model. This demonstrates the contribution of the MoE model to COMET's performance. Furthermore, the gate loss, EWC loss, and SL synergistically enhance the effectiveness of the MoE model.

\begin{figure}[h]
\centering
\includegraphics[width=0.3\linewidth]{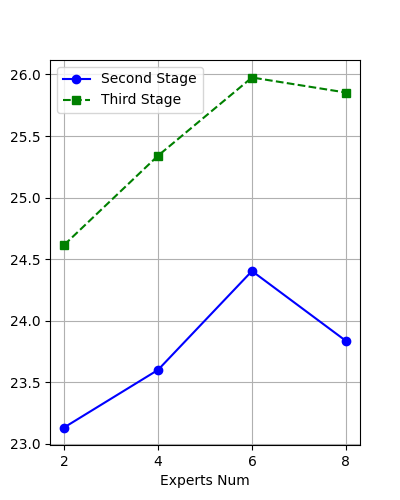}
\caption{Ablation on the number of experts}
\label{fig:expert_ablation}
\vspace{-2mm}
\end{figure}

\textbf{Effectiveness of the number of experts:}
Figure \ref{fig:expert_ablation} presents the ablation study on the number of experts, revealing that as the number of experts increases, the performance on the V-A task improves significantly. However, when the number of experts exceeds a certain threshold (6 in our experiments), the task performance begins to degrade, exhibiting similar trends in both the Second Stage and Third Stage. We hypothesize that this decline is due to overfitting. Consequently, we selected six experts for our experiments.

\subsection{Qualitative Analysis}
\begin{figure}[h]
\center
  \includegraphics[width=14cm]{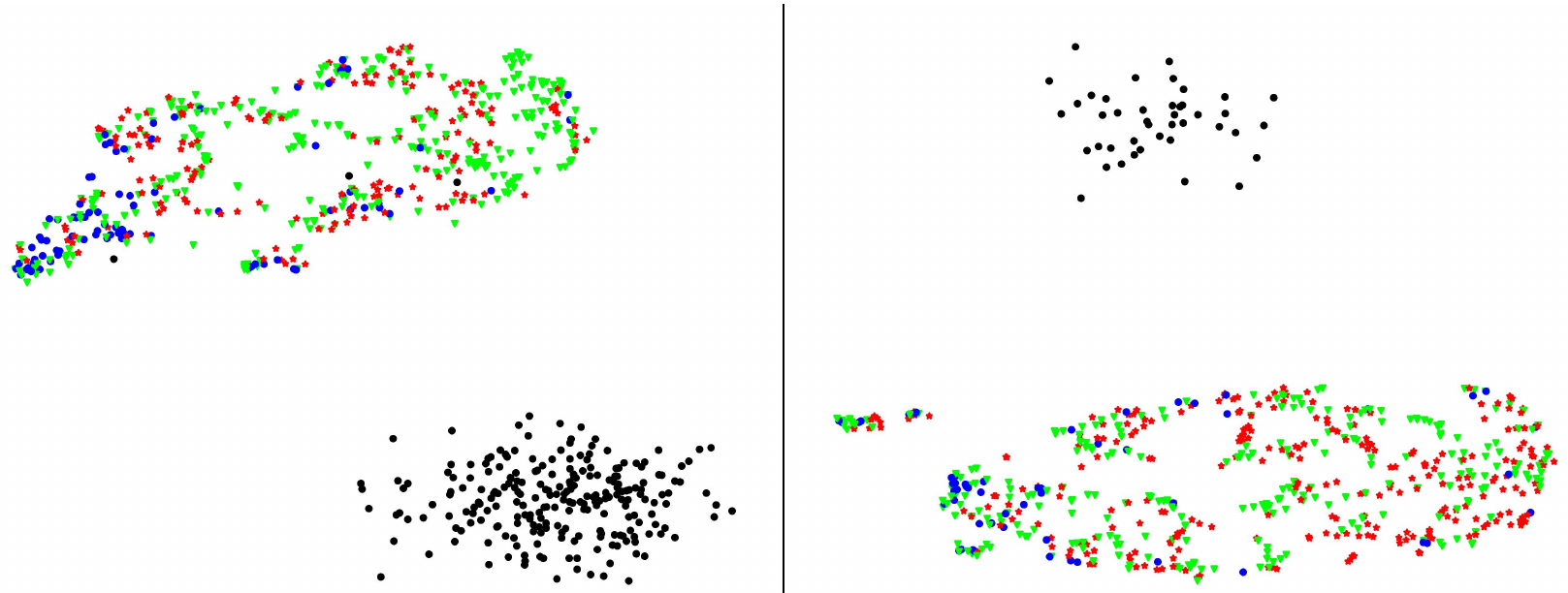}
  \caption{Visualization of discrete codes. Red indicates codes effectively activated by all three modalities (video, audio, and text), green represents codes effectively activated by two modalities, blue denotes codes effectively activated by one modality, and black signifies codes that are not effectively activated. }
  % A modality is considered to effectively activate a code if the activation count from that modality constitutes 10\% or more of the total activation counts.}
  \label{fig:distribution}
  \vspace{-2mm}
\end{figure}

Figure \ref{fig:distribution} visualizes the discrete codes generated by DCID and DCID+COMET. We used the VALOR32K~\cite{chen2023valor} dataset, which includes video, audio, and text tri-modal paired data, to quantize the activations of the second-stage models trained with DCID and DCID+COMET. 

From the observations, it is evident that DCID has a significant number of codes that are not effectively activated. Moreover, compared to DCID+COMET, DCID has more codes that are activated by a single modality and fewer codes that are activated by all three modalities. In contrast, the majority of codes in DCID+COMET are activated by at least two modalities, demonstrating that COMET effectively enhances the unified representation across multiple modalities and improves the utilization rate of the codebook.

\section{Conclusion}
In conclusion, our work addresses the intrinsic limitations of the previous Cross-Modal Generalization (CMG) framework and propose a new framework named COMET. Through the introduction of a Continual Mixture of Experts Adapter (CMoE-Adapter) and a pseudo-modality replay mechanism (PMR), we enable the model to map different modalities into a unified semantic space while maintaining the alignment of original modalities. The CMoE-Adapter enhances the model's encoding capacity and preserves previously learned knowledge using adaptive Elastic Weight Consolidation (EWC) loss. Meanwhile, the PMR mechanism employs a dynamic expansion codebook and pseudo-modality sequences to ensure that new modalities are effectively integrated into the existing semantic space. These contributions significantly advance the field of unified multimodal representation, allowing for more scalable and flexible learning from diverse multimodal data.

\bibliographystyle{unsrt}
\bibliography{neurips_2024}

\begin{thebibliography}{10}

\bibitem{girdhar2023imagebind}
Rohit Girdhar, Alaaeldin El-Nouby, Zhuang Liu, Mannat Singh, Kalyan~Vasudev Alwala, Armand Joulin, and Ishan Misra.
\newblock Imagebind: One embedding space to bind them all.
\newblock In {\em Proceedings of the IEEE/CVF Conference on Computer Vision and Pattern Recognition}, pages 15180--15190, 2023.

\bibitem{radford2021learning}
Alec Radford, Jong~Wook Kim, Chris Hallacy, Aditya Ramesh, Gabriel Goh, Sandhini Agarwal, Girish Sastry, Amanda Askell, Pamela Mishkin, Jack Clark, et~al.
\newblock Learning transferable visual models from natural language supervision.
\newblock In {\em International conference on machine learning}, pages 8748--8763. PMLR, 2021.

\bibitem{liang2022mind}
Victor~Weixin Liang, Yuhui Zhang, Yongchan Kwon, Serena Yeung, and James~Y Zou.
\newblock Mind the gap: Understanding the modality gap in multi-modal contrastive representation learning.
\newblock {\em Advances in Neural Information Processing Systems}, 35:17612--17625, 2022.

\bibitem{zhao2022towards}
Yang Zhao, Chen Zhang, Haifeng Huang, Haoyuan Li, and Zhou Zhao.
\newblock Towards effective multi-modal interchanges in zero-resource sounding object localization.
\newblock {\em Advances in Neural Information Processing Systems}, 35:38089--38102, 2022.

\bibitem{akbari2021vatt}
Hassan Akbari, Liangzhe Yuan, Rui Qian, Wei-Hong Chuang, Shih-Fu Chang, Yin Cui, and Boqing Gong.
\newblock Vatt: Transformers for multimodal self-supervised learning from raw video, audio and text.
\newblock {\em Advances in Neural Information Processing Systems}, 34:24206--24221, 2021.

\bibitem{you2022learning}
Haoxuan You, Luowei Zhou, Bin Xiao, Noel Codella, Yu~Cheng, Ruochen Xu, Shih-Fu Chang, and Lu~Yuan.
\newblock Learning visual representation from modality-shared contrastive language-image pre-training.
\newblock In {\em Computer Vision--ECCV 2022: 17th European Conference, Tel Aviv, Israel, October 23--27, 2022, Proceedings, Part XXVII}, pages 69--87. Springer, 2022.

\bibitem{liu2021cross}
Alexander~H Liu, SouYoung Jin, Cheng-I~Jeff Lai, Andrew Rouditchenko, Aude Oliva, and James Glass.
\newblock Cross-modal discrete representation learning.
\newblock {\em arXiv preprint arXiv:2106.05438}, 2021.

\bibitem{duan2022multi}
Jiali Duan, Liqun Chen, Son Tran, Jinyu Yang, Yi~Xu, Belinda Zeng, and Trishul Chilimbi.
\newblock Multi-modal alignment using representation codebook.
\newblock In {\em Proceedings of the IEEE/CVF Conference on Computer Vision and Pattern Recognition}, pages 15651--15660, 2022.

\bibitem{van2017neural}
Aaron Van Den~Oord, Oriol Vinyals, et~al.
\newblock Neural discrete representation learning.
\newblock {\em Advances in neural information processing systems}, 30, 2017.

\bibitem{ji2024wavtokenizer}
Shengpeng Ji, Ziyue Jiang, Wen Wang, Yifu Chen, Minghui Fang, Jialong Zuo, Qian Yang, Xize Cheng, Zehan Wang, Ruiqi Li, et~al.
\newblock Wavtokenizer: an efficient acoustic discrete codec tokenizer for audio language modeling.
\newblock {\em arXiv preprint arXiv:2408.16532}, 2024.

\bibitem{xia2024achieving}
Yan Xia, Hai Huang, Jieming Zhu, and Zhou Zhao.
\newblock Achieving cross modal generalization with multimodal unified representation.
\newblock {\em Advances in Neural Information Processing Systems}, 36, 2024.

\bibitem{wang2019vatex}
Xin Wang, Jiawei Wu, Junkun Chen, Lei Li, Yuan-Fang Wang, and William~Yang Wang.
\newblock Vatex: A large-scale, high-quality multilingual dataset for video-and-language research.
\newblock In {\em Proceedings of the IEEE/CVF International Conference on Computer Vision}, pages 4581--4591, 2019.

\bibitem{kim2019audiocaps}
Chris~Dongjoo Kim, Byeongchang Kim, Hyunmin Lee, and Gunhee Kim.
\newblock Audiocaps: Generating captions for audios in the wild.
\newblock In {\em Proceedings of the 2019 Conference of the North American Chapter of the Association for Computational Linguistics: Human Language Technologies, Volume 1 (Long and Short Papers)}, pages 119--132, 2019.

\bibitem{young2014image}
Peter Young, Alice Lai, Micah Hodosh, and Julia Hockenmaier.
\newblock From image descriptions to visual denotations: New similarity metrics for semantic inference over event descriptions.
\newblock {\em Transactions of the Association for Computational Linguistics}, 2:67--78, 2014.

\bibitem{zen2019libritts}
Heiga Zen, Viet Dang, Rob Clark, Yu~Zhang, Ron~J Weiss, Ye~Jia, Zhifeng Chen, and Yonghui Wu.
\newblock Libritts: A corpus derived from librispeech for text-to-speech.
\newblock {\em arXiv preprint arXiv:1904.02882}, 2019.

\bibitem{seo2020urvos}
Seonguk Seo, Joon-Young Lee, and Bohyung Han.
\newblock Urvos: Unified referring video object segmentation network with a large-scale benchmark.
\newblock In {\em Computer Vision--ECCV 2020: 16th European Conference, Glasgow, UK, August 23--28, 2020, Proceedings, Part XV 16}, pages 208--223. Springer, 2020.

\bibitem{avel}
Yapeng Tian, Jing Shi, Bochen Li, Zhiyao Duan, and Chenliang Xu.
\newblock Audio-visual event localization in unconstrained videos.
\newblock In {\em Proceedings of the European Conference on Computer Vision (ECCV)}, pages 247--263, 2018.

\bibitem{tian2020unified}
Yapeng Tian, Dingzeyu Li, and Chenliang Xu.
\newblock Unified multisensory perception: Weakly-supervised audio-visual video parsing.
\newblock In {\em Computer Vision--ECCV 2020: 16th European Conference, Glasgow, UK, August 23--28, 2020, Proceedings, Part III 16}, pages 436--454. Springer, 2020.

\bibitem{chen2023valor}
Sihan Chen, Xingjian He, Longteng Guo, Xinxin Zhu, Weining Wang, Jinhui Tang, and Jing Liu.
\newblock Valor: Vision-audio-language omni-perception pretraining model and dataset.
\newblock {\em arXiv preprint arXiv:2304.08345}, 2023.

\bibitem{petridis2018audio}
Stavros Petridis, Themos Stafylakis, Pingchuan Ma, Georgios Tzimiropoulos, and Maja Pantic.
\newblock Audio-visual speech recognition with a hybrid ctc/attention architecture.
\newblock In {\em 2018 IEEE Spoken Language Technology Workshop (SLT)}, pages 513--520. IEEE, 2018.

\bibitem{andonian2022robust}
Alex Andonian, Shixing Chen, and Raffay Hamid.
\newblock Robust cross-modal representation learning with progressive self-distillation.
\newblock In {\em Proceedings of the IEEE/CVF Conference on Computer Vision and Pattern Recognition}, pages 16430--16441, 2022.

\bibitem{chen2020uniter}
Yen-Chun Chen, Linjie Li, Licheng Yu, Ahmed El~Kholy, Faisal Ahmed, Zhe Gan, Yu~Cheng, and Jingjing Liu.
\newblock Uniter: Universal image-text representation learning.
\newblock In {\em European conference on computer vision}, pages 104--120. Springer, 2020.

\bibitem{wang2025towards}
Shulei Wang, Wang Lin, Hai Huang, Hanting Wang, Sihang Cai, WenKang Han, Tao Jin, Jingyuan Chen, Jiacheng Sun, Jieming Zhu, et~al.
\newblock Towards transformer-based aligned generation with self-coherence guidance.
\newblock {\em arXiv preprint arXiv:2503.17675}, 2025.

\bibitem{lu2022unified}
Jiasen Lu, Christopher Clark, Rowan Zellers, Roozbeh Mottaghi, and Aniruddha Kembhavi.
\newblock Unified-io: A unified model for vision, language, and multi-modal tasks.
\newblock In {\em The Eleventh International Conference on Learning Representations}, 2022.

\bibitem{hdcid}
Hai Huang, Yan Xia, Shengpeng Ji, Shulei Wang, Hanting Wang, Jieming Zhu, Zhenhua Dong, and Zhou Zhao.
\newblock Unlocking the potential of multimodal unified discrete representation through training-free codebook optimization and hierarchical alignment.
\newblock {\em arXiv preprint arXiv:2403.05168}, 2024.

\bibitem{huang2025semantic}
Hai Huang, Shulei Wang, and Yan Xia.
\newblock Semantic residual for multimodal unified discrete representation.
\newblock In {\em ICASSP 2025-2025 IEEE International Conference on Acoustics, Speech and Signal Processing (ICASSP)}, pages 1--5. IEEE, 2025.

\bibitem{sarkar2024xkd}
Pritam Sarkar and Ali Etemad.
\newblock Xkd: Cross-modal knowledge distillation with domain alignment for video representation learning.
\newblock In {\em Proceedings of the AAAI Conference on Artificial Intelligence}, volume~38, pages 14875--14885, 2024.

\bibitem{wang2022vlmixer}
Teng Wang, Wenhao Jiang, Zhichao Lu, Feng Zheng, Ran Cheng, Chengguo Yin, and Ping Luo.
\newblock Vlmixer: Unpaired vision-language pre-training via cross-modal cutmix.
\newblock In {\em International Conference on Machine Learning}, pages 22680--22690. PMLR, 2022.

\bibitem{han2021learning}
Chi Han, Mingxuan Wang, Heng Ji, and Lei Li.
\newblock Learning shared semantic space for speech-to-text translation.
\newblock {\em arXiv preprint arXiv:2105.03095}, 2021.

\bibitem{NEURIPS2023_c89f0984}
Yan Xia, Hai Huang, Jieming Zhu, and Zhou Zhao.
\newblock Achieving cross modal generalization with multimodal unified representation.
\newblock In A.~Oh, T.~Neumann, A.~Globerson, K.~Saenko, M.~Hardt, and S.~Levine, editors, {\em Advances in Neural Information Processing Systems}, volume~36, pages 63529--63541. Curran Associates, Inc., 2023.

\bibitem{li2017learning}
Zhizhong Li and Derek Hoiem.
\newblock Learning without forgetting.
\newblock {\em IEEE transactions on pattern analysis and machine intelligence}, 40(12):2935--2947, 2017.

\bibitem{cermelli2020modeling}
Fabio Cermelli, Massimiliano Mancini, Samuel~Rota Bulo, Elisa Ricci, and Barbara Caputo.
\newblock Modeling the background for incremental learning in semantic segmentation.
\newblock In {\em Proceedings of the IEEE/CVF Conference on Computer Vision and Pattern Recognition}, pages 9233--9242, 2020.

\bibitem{kanakis2020reparameterizing}
Menelaos Kanakis, David Bruggemann, Suman Saha, Stamatios Georgoulis, Anton Obukhov, and Luc Van~Gool.
\newblock Reparameterizing convolutions for incremental multi-task learning without task interference.
\newblock In {\em Computer Vision--ECCV 2020: 16th European Conference, Glasgow, UK, August 23--28, 2020, Proceedings, Part XX 16}, pages 689--707. Springer, 2020.

\bibitem{wallingford2022task}
Matthew Wallingford, Hao Li, Alessandro Achille, Avinash Ravichandran, Charless Fowlkes, Rahul Bhotika, and Stefano Soatto.
\newblock Task adaptive parameter sharing for multi-task learning.
\newblock In {\em Proceedings of the IEEE/CVF Conference on Computer Vision and Pattern Recognition}, pages 7561--7570, 2022.

\bibitem{tasar2019incremental}
Onur Tasar, Yuliya Tarabalka, and Pierre Alliez.
\newblock Incremental learning for semantic segmentation of large-scale remote sensing data.
\newblock {\em IEEE Journal of Selected Topics in Applied Earth Observations and Remote Sensing}, 12(9):3524--3537, 2019.

\bibitem{saporta2022multi}
Antoine Saporta, Arthur Douillard, Tuan-Hung Vu, Patrick P{\'e}rez, and Matthieu Cord.
\newblock Multi-head distillation for continual unsupervised domain adaptation in semantic segmentation.
\newblock In {\em Proceedings of the IEEE/CVF Conference on Computer Vision and Pattern Recognition}, pages 3751--3760, 2022.

\bibitem{peng2021adaptive}
Wei Peng, Xiaopeng Hong, and Guoying Zhao.
\newblock Adaptive modality distillation for separable multimodal sentiment analysis.
\newblock {\em IEEE Intelligent Systems}, 36(3):82--89, 2021.

\bibitem{zhang2023vqacl}
Xi~Zhang, Feifei Zhang, and Changsheng Xu.
\newblock Vqacl: A novel visual question answering continual learning setting.
\newblock In {\em Proceedings of the IEEE/CVF Conference on Computer Vision and Pattern Recognition}, pages 19102--19112, 2023.

\bibitem{isele2018selective}
David Isele and Akansel Cosgun.
\newblock Selective experience replay for lifelong learning.
\newblock In {\em Proceedings of the AAAI Conference on Artificial Intelligence}, volume~32, 2018.

\bibitem{lavda2018continual}
Frantzeska Lavda, Jason Ramapuram, Magda Gregorova, and Alexandros Kalousis.
\newblock Continual classification learning using generative models.
\newblock {\em arXiv preprint arXiv:1810.10612}, 2018.

\bibitem{lopez2017gradient}
David Lopez-Paz and Marc'Aurelio Ranzato.
\newblock Gradient episodic memory for continual learning.
\newblock {\em Advances in neural information processing systems}, 30, 2017.

\bibitem{kirkpatrick2017overcoming}
James Kirkpatrick, Razvan Pascanu, Neil Rabinowitz, Joel Veness, Guillaume Desjardins, Andrei~A Rusu, Kieran Milan, John Quan, Tiago Ramalho, Agnieszka Grabska-Barwinska, et~al.
\newblock Overcoming catastrophic forgetting in neural networks.
\newblock {\em Proceedings of the national academy of sciences}, 114(13):3521--3526, 2017.

\bibitem{zenke2017continual}
Friedemann Zenke, Ben Poole, and Surya Ganguli.
\newblock Continual learning through synaptic intelligence.
\newblock In {\em International conference on machine learning}, pages 3987--3995. PMLR, 2017.

\bibitem{yan2021dynamically}
Shipeng Yan, Jiangwei Xie, and Xuming He.
\newblock Der: Dynamically expandable representation for class incremental learning.
\newblock In {\em Proceedings of the IEEE/CVF conference on computer vision and pattern recognition}, pages 3014--3023, 2021.

\bibitem{ye2023self}
Fei Ye and Adrian~G Bors.
\newblock Self-evolved dynamic expansion model for task-free continual learning.
\newblock In {\em Proceedings of the IEEE/CVF International Conference on Computer Vision}, pages 22102--22112, 2023.

\bibitem{yu2024boosting}
Jiazuo Yu, Yunzhi Zhuge, Lu~Zhang, Dong Wang, Huchuan Lu, and You He.
\newblock Boosting continual learning of vision-language models via mixture-of-experts adapters.
\newblock {\em arXiv preprint arXiv:2403.11549}, 2024.

\bibitem{chen2023lifelong}
Wuyang Chen, Yanqi Zhou, Nan Du, Yanping Huang, James Laudon, Zhifeng Chen, and Claire Cui.
\newblock Lifelong language pretraining with distribution-specialized experts.
\newblock In {\em International Conference on Machine Learning}, pages 5383--5395. PMLR, 2023.

\bibitem{pan2022wnet}
Wenwen Pan, Haonan Shi, Zhou Zhao, Jieming Zhu, Xiuqiang He, Zhigeng Pan, Lianli Gao, Jun Yu, Fei Wu, and Qi~Tian.
\newblock Wnet: Audio-guided video object segmentation via wavelet-based cross-modal denoising networks.
\newblock In {\em Proceedings of the IEEE/CVF Conference on Computer Vision and Pattern Recognition}, pages 1320--1331, 2022.

\bibitem{wu2022language}
Jiannan Wu, Yi~Jiang, Peize Sun, Zehuan Yuan, and Ping Luo.
\newblock Language as queries for referring video object segmentation.
\newblock In {\em Proceedings of the IEEE/CVF Conference on Computer Vision and Pattern Recognition}, pages 4974--4984, 2022.

\bibitem{chen2011collecting}
David Chen and William~B Dolan.
\newblock Collecting highly parallel data for paraphrase evaluation.
\newblock In {\em Proceedings of the 49th annual meeting of the association for computational linguistics: human language technologies}, pages 190--200, 2011.

\bibitem{test1k}
Youngjae Yu, Jongseok Kim, and Gunhee Kim.
\newblock A joint sequence fusion model for video question answering and retrieval.
\newblock In {\em Proceedings of the European conference on computer vision (ECCV)}, pages 471--487, 2018.

\bibitem{msrvtt}
Jun Xu, Tao Mei, Ting Yao, and Yong Rui.
\newblock Msr-vtt: A large video description dataset for bridging video and language.
\newblock In {\em Proceedings of the IEEE conference on computer vision and pattern recognition}, pages 5288--5296, 2016.

\bibitem{senocak2018learning}
Arda Senocak, Tae-Hyun Oh, Junsik Kim, Ming-Hsuan Yang, and In~So Kweon.
\newblock Learning to localize sound source in visual scenes.
\newblock In {\em Proceedings of the IEEE Conference on Computer Vision and Pattern Recognition}, pages 4358--4366, 2018.

\bibitem{drossos2020clotho}
Konstantinos Drossos, Samuel Lipping, and Tuomas Virtanen.
\newblock Clotho: An audio captioning dataset.
\newblock In {\em ICASSP 2020-2020 IEEE International Conference on Acoustics, Speech and Signal Processing (ICASSP)}, pages 736--740. IEEE, 2020.

\bibitem{lin2014microsoft}
Tsung-Yi Lin, Michael Maire, Serge Belongie, James Hays, Pietro Perona, Deva Ramanan, Piotr Doll{\'a}r, and C~Lawrence Zitnick.
\newblock Microsoft coco: Common objects in context.
\newblock In {\em Computer Vision--ECCV 2014: 13th European Conference, Zurich, Switzerland, September 6-12, 2014, Proceedings, Part V 13}, pages 740--755. Springer, 2014.

\end{thebibliography}

\appendix

\section{Related Work}

\textbf{Multi-Modal Unified Representation}

Multi-modal unified representation has emerged as a promising research field in recent years, which aims to align different modalities into a shared latent space, and can be allocated into two parts: implicit unified representation \cite{petridis2018audio, andonian2022robust, chen2020uniter, wang2025towards} and explicit unified representation \cite{zhao2022towards, lu2022unified, liu2021cross, hdcid, huang2025semantic}. The former is dedicated to leveraging contrastive learning to bring different modalities closer in the latent space \cite{sarkar2024xkd, andonian2022robust}, or using a general modality-agnostic Encoder to encode various modalities \cite{wang2022vlmixer}; whereas the latter employs methods such as optimal transport \cite{duan2022multi}, vector quantization \cite{liu2021cross}, etc., to map information from different modalities onto a set of universal prototypes or dictionary vectors. These approaches have been applied across various combinations of modalities, such as image-text \cite{duan2022multi}, audio-video \cite{liu2021cross}, and speech-text domains \cite{han2021learning}. Despite the significant achievements of the abovementioned methods, most of these works perform coarse-grained unification of representations, lacking fine-grained alignment between different modalities. Recently, Xia et al. proposed a new task named Cross Modal Generalization (CMG) \cite{NEURIPS2023_c89f0984}, which aims to learn fine-grained unified representation from paired multimodal data during pre-training, and then achieve zero-shot transfer ability from the labeled modality to other unlabeled modalities in downstream tasks. Their method requires large-scale paired modalities during the pretraining stage, however, obtaining such extensive fully paired multimodal data becomes challenging when attempting to achieve a unified representation of three or more modalities.
%因此在本文中，我们提出了一个增强版本的CMG，可以在预训练阶段利用持续学习的方式不断的扩充两两配对的数据，极大的缓解了预训练数据不足的问题。

\textbf{Continue Learning with Mixture of Experts}

The goal of continual learning is to adapt to new information over time without forgetting previously acquired knowledge. This process mimics the human ability to accumulate and build upon knowledge throughout life. Existing methods can be divided into three main applications, Class Incremental Learning \cite{li2017learning, cermelli2020modeling}, Task Incremental Learning \cite{kanakis2020reparameterizing, wallingford2022task}, Domain Incremental Learning \cite{tasar2019incremental, saporta2022multi} and Modality Incremental Learning \cite{peng2021adaptive, zhang2023vqacl}. The main technologies currently encompass the following categories: memory-based \cite{isele2018selective, lavda2018continual, lopez2017gradient}, regularization-based \cite{kirkpatrick2017overcoming, zenke2017continual}, and dynamic-based models \cite{yan2021dynamically, ye2023self}. Recently, Mixture of Experts (MoE) has been adopted into Incremental Learning, which can utilize different experts for different scenarios. \cite{yu2024boosting, chen2023lifelong} introduce a dynamic expansion MoE-Adapters framework that can access shareable knowledge from frozen experts trained in historical tasks and optimize unfrozen experts to acquire novel knowledge from new tasks.

\section{Downstream Task}
\label{sec:appendix_downstream}
\textbf{Cross-modal Event Classification on AVE Dataset:}~\cite{avel} We train on one modality (e.g., video) and evaluate on another (e.g., audio). \\
\textbf{Cross-modal Event Localization on AVVP Dataset:}~\cite{tian2020unified} Localizing events in one modality and transferring to the other. \\
\textbf{Cross-dataset Localization/Classification:} Training on classification in AVE and evaluating localization in AVVP, transferring across datasets. \\
\textbf{Cross-modal Video Segmentation on Ref-Youtube-VOS Dataset:}~\cite{seo2020urvos} The training is for segmenting video based on a text query, while testing involves segmenting video based on the corresponding speech query of the text. Though Ref-Youtube-VOS lacks speech data, this approach uses data adapted from AVOS~\cite{pan2022wnet} by combining datasets. The downstream model employs ReferFormer~\cite{wu2022language}. \textbf{Cross-modal Zero-shot Retrieval:} 
\textbf{Video-Text:} MSVD~\cite{chen2011collecting}; we adopt a process similar to the test set~\cite{test1k} consists of 1,000 video-text pairs from MSRVTT~\cite{msrvtt}. This task tests the retrieval effectiveness between videos and text.\\
\textbf{Image-Audio:} Flickr Sound~\cite{senocak2018learning}; evaluates the alignment effectiveness between images and audio without prior direct training on paired samples. \\
\textbf{Audio-Text:} Clotho~\cite{drossos2020clotho}; assesses the zero-shot retrieval capability for audio-text alignment. \\
\textbf{Image-Text:} MS-COCO~\cite{lin2014microsoft}; focuses on the alignment performance between images and text in a zero-shot retrieval context, Using a test subset with a sample size of 1000.

\section{Seen data pairs experiments supplement}
Table \ref{tab:unseen_IA} shows effects similar to those in Table\ref{tab:unseen_va}, where COMET effectively helped improve the performance of various models, and remained effective even after changing the order of pre-training pairs.

\label{sec:appendix_seen}
\begin{table}[h]
\caption{Retrieval results in image-audio pairs. $\dagger$ indicates pre-training in the order of AT-VT-IT and the other settings in the order of VT-AT-IT.}
\label{tab:unseen_IA}
\centering
\begin{tabular}{cccccccc}
\toprule
\centering
\multirow{3}{*}{Method} & \multicolumn{7}{c}{Third Stage} \\
\cmidrule{2-8}
& \multicolumn{3}{c}{I $\rightarrow$ A} & \multicolumn{3}{c}{A $\rightarrow$ I} & \multirow{2}{*}{Avg.} \\
& R@1 & R@5 & R10 & R@1 & R@5 & R10 &\\
\midrule
TURN~\cite{zhao2022towards} & 0.4 & 3.6 & 6.8 & 1.0 & 3.4 & 5.2 & 3.4 \\
CMCM~\cite{liu2021cross} & 0.6 & 4.0 & 7.6 & 0.8 & 3.8 & 6.8 & 3.93 \\
DCID~\cite{xia2024achieving} & \textbf{1.2} & \textbf{4.4} & 7.4 & \textbf{1.4} & 4.0 & 5.8 & 4.03 \\
TURN~\cite{zhao2022towards}+COMET & 0.8 & 3.2 & 7.4 & 1.2 & 4.0 & 5.6 & 3.7 \\
CMCM~\cite{liu2021cross}+COMET & \textbf{1.2} & 4.2 & 8.2 & 1.2 &\textbf{4.6} & 7.6 & 4.5\\
DCID~\cite{xia2024achieving}+COMET & \textbf{1.2} & 4.2 & \textbf{9.0} & 0.4 & \textbf{4.6}& \textbf{8.6} & \textbf{4.67} \\
\midrule
DCID~\cite{xia2024achieving} $\dagger$ & 0.6 & 2.2 & 5.2 & \textbf{0.6} & 2.2 & 5.2 & 2.67 \\
DCID~\cite{xia2024achieving}+COMET $\dagger$ & \textbf{1.2} & \textbf{4.2} & \textbf{7.0} & 0.4 & \textbf{2.8} & \textbf{6.6} & \textbf{3.7} \\
\bottomrule
\end{tabular}%
\end{table}

\section{Unseen data pairs experiments supplement}
Table \ref{tab:ablation_ia} presents a slightly different scenario. While DCID slightly outperforms some 'partial module removal' methods in certain tasks, the performance improvement from using individual modules alone is not significant. However, when multiple modules are employed simultaneously, they exhibit synergistic effects, collectively enhancing the model's performance across various tasks.

% \label{sec:appendix_unseen}
% \begin{table}[h]
% \caption{Ablation studies of image-audio in the third training stage.(pre-trained on VT-AT-IT)}
% \label{tab:ablation_ia}
% \resizebox{\textwidth}{!}
% {%
% \begin{tabular}{cccccccccccccc}
% \toprule
% \centering
% \multirow{3}{*}{PM} & \multirow{3}{*}{MOE} & \multirow{3}{*}{\mathcal{L}_{Gate}} & \multirow{3}{*}{\mathcal{L}_{EWC}} & \multirow{3}{*}{\mathcal{SL}} & \multicolumn{9}{c}{Third Stage}    \\   \cmidrule{6-14}
% & & & & & \multicolumn{3}{c}{I $\rightarrow$ A} & & \multicolumn{3}{c}{A $\rightarrow$ I} & &\multirow{2}{*}{\multicolumn{1}{c}{Avg.}} \\
% & & & & & R@1 & R@5 & R10 & & R@1 & R@5 & R10 & & \\
% \midrule
% - & \checkmark & \checkmark & \checkmark & \checkmark & 0.6 & 3.6 & 7.2 & & 1.0 & 3.6 & 7.0 &  & 3.83 \\
% \checkmark & - & - & - & - & 0.8 & 3.6 & 8.2 & & 0.6 & 3.2 & 6.4 &  & 3.8 \\
% \checkmark & \checkmark & - & \checkmark & \checkmark & 0.4 & \textbf{5.0} & \textbf{10.2} & & 0.2 & 3.2 & 8.0 &  & 4.5 \\
% \checkmark & \checkmark & \checkmark & - & \checkmark & 0.6 & 4.0 & 7.4 & & 1.2 & 3.6 & 6.8 &  & 3.93 \\
% \checkmark & \checkmark & \checkmark & \checkmark& - & 0.8 & 4.0 & 7.6 & & 1.2 & 3.4 & 6.8 &  & 3.97 \\
% - & - & - & - & -& \textbf{1.2} & 4.4 & 7.4 & & \textbf{1.4} & 4.0 & 5.8 &  & 4.03 \\
% \checkmark & \checkmark & \checkmark & \checkmark & \checkmark & \textbf{1.2} & 4.2 & 9.0 & & 0.4 & \textbf{4.6} & \textbf{8.6} &  & \textbf{4.66} \\
% \bottomrule
% \end{tabular}%
% }
% \end{table}

\label{sec:appendix_unseen}
\begin{table}[h]
\caption{Ablation studies of image-audio in the third training stage.(pre-trained on VT-AT-IT)}
\label{tab:ablation_ia}
\centering % 将 \centering 移到表格外部
\resizebox{\textwidth}{!}{%
\begin{tabular}{cccccccccccccc}
\toprule
\multirow{3}{*}{PM} & \multirow{3}{*}{MOE} & \multirow{3}{*}{$\mathcal{L}_{Gate}$} & \multirow{3}{*}{$\mathcal{L}_{EWC}$} & \multirow{3}{*}{$\mathcal{SL}$} & \multicolumn{9}{c}{Third Stage} \\  
\cmidrule(lr){6-14}
& & & & & \multicolumn{3}{c}{I $\rightarrow$ A} & \multicolumn{1}{c}{} & \multicolumn{3}{c}{A $\rightarrow$ I} & \multicolumn{1}{c}{} & \multirow{2}{*}{Avg.} \\
& & & & & R@1 & R@5 & R10 & & R@1 & R@5 & R10 & & \\
\midrule
- & $\checkmark$ & $\checkmark$ & $\checkmark$ & $\checkmark$ & 0.6 & 3.6 & 7.2 & & 1.0 & 3.6 & 7.0 & & 3.83 \\
$\checkmark$ & - & - & - & - & 0.8 & 3.6 & 8.2 & & 0.6 & 3.2 & 6.4 & & 3.8 \\
$\checkmark$ & $\checkmark$ & - & $\checkmark$ & $\checkmark$ & 0.4 & \textbf{5.0} & \textbf{10.2} & & 0.2 & 3.2 & 8.0 & & 4.5 \\
$\checkmark$ & $\checkmark$ & $\checkmark$ & - & $\checkmark$ & 0.6 & 4.0 & 7.4 & & 1.2 & 3.6 & 6.8 & & 3.93 \\
$\checkmark$ & $\checkmark$ & $\checkmark$ & $\checkmark$ & - & 0.8 & 4.0 & 7.6 & & 1.2 & 3.4 & 6.8 & & 3.97 \\
- & - & - & - & - & \textbf{1.2} & 4.4 & 7.4 & & \textbf{1.4} & 4.0 & 5.8 & & 4.03 \\
$\checkmark$ & $\checkmark$ & $\checkmark$ & $\checkmark$ & $\checkmark$ & \textbf{1.2} & 4.2 & 9.0 & & 0.4 & \textbf{4.6} & \textbf{8.6} & & \textbf{4.66} \\
\bottomrule
\end{tabular}%
}
\end{table}

\section{Limitation}
\label{sec:appendix_limit}
Insufficient pre-training can be addressed by adding more pre-training data to enhance the persuasiveness of the experiment. Additionally, richer data pairings and more training phases can also help improve the experiment's persuasiveness.

\section{Compute Resources}
\label{sec:appendix_compute_resource}
Using a single Nvidia RTX 3090TI as an example, pre-training for VT-AT-IT and AT-VT-IT requires 18 hours, while VT-ST-AT pre-training takes 13 hours. Downstream V$\leftrightarrow$A classification and localization can be completed within 20 minutes, and other cross-modal zero-shot retrieval tasks require only 5 minutes each. The semantic segmentation task involving text$\rightarrow$speech, due to its high complexity and the use of the ReferFormer~\cite{wu2022language} model, requires 72 hours for training and 1 hour for inference.

\end{document}